\documentclass[letterpaper, 10 pt, journal, twoside]{ieeetran}
\usepackage{caption}
\usepackage{amsmath,amsfonts,amssymb}
\usepackage{algorithmic}
\usepackage{algorithm}
\usepackage{array}
\usepackage[caption=false,font=normalsize,labelfont=sf,textfont=sf]{subfig}
\usepackage{textcomp}
\usepackage{stfloats}
\usepackage{url}
\usepackage{verbatim}
\usepackage{graphicx}
\usepackage{cite}
\usepackage{authblk}
\usepackage{xcolor}
\usepackage{wrapfig}

\usepackage{bm}          
\usepackage{booktabs}
\usepackage{siunitx}
\usepackage{hyperref}
\usepackage{mathtools}

\hypersetup{
    colorlinks = true,
    linkcolor  = blue!60!black,
    citecolor  = green!50!black,
    urlcolor   = blue!70!black
}

\graphicspath{{figures/}}

\newcommand{\R}{\mathbb{R}}
\newcommand{\vect}[1]{\bm{#1}}
\newcommand{\mat}[1]{\bm{#1}}
\newcommand{\dq}{\dot{q}}

\DeclareMathOperator*{\argmin}{arg\,min}

\begin{document}

\title{Identification of a Physics-Based Electrical Power Consumption Model for the Unitree G1 Humanoid Arm}

\author{}
\author{Nestor N. Deniz$^{1}$, Sebastian Vega$^{1}$, Simon Parsons$^{2}$ and Fernando A. Auat Cheein$^{1}$
\thanks{} 
\thanks{$^{1}$Nestor N. Deniz, Sebastian Vega and Fernando Auat Cheein are with the Engineering Department at Harper Adams University, Newport, Shropshire TF10 8NB, UK.
        {\tt\footnotesize ndeniz@harper-adams.ac.uk}}%
\thanks{$^{2} $Simon Parsons is with Lincoln Institute for Agri-Food Technology and Lincoln Centre for Autonomous Systems.}%
}

\maketitle

\begin{abstract}
Accurate prediction of electrical power consumption is essential for
energy-aware motion planning, battery management, and thermal monitoring in
battery-powered humanoid robots. This letter presents a physics-based,
linear-in-parameters model for the electrical power consumption of the
seven-degree-of-freedom left arm of the Unitree~G1 humanoid robot. The proposed
formulation combines actuator loss terms with a baseline-torque correction that
captures changes in gravity-compensation load and enables accurate prediction
of negative net power trajectories. Pairwise interaction terms are introduced
to model power coupling during simultaneous multi-joint motion.
Model parameters are identified from experimental data collected on a physical
Unitree~G1 using onboard power measurements as the regression target. Across
897 trajectories covering single-joint and coordinated arm motions at multiple
speed levels, the identified model achieves $R^2 = 0.933$ with an
RMSE of 1.07 (W). Validation on 46 trajectories executed at
previously unseen speeds yields $R^2 = 0.965$, demonstrating strong
generalisation beyond the identification dataset. Analysis of the identified
parameters reveals distinct power-consumption characteristics across the arm,
with viscous friction dominating most joints (shoulder pitch and all three
wrist joints), copper losses dominating shoulder yaw and the elbow, and
shoulder roll uniquely dominated by Coulomb friction.
\end{abstract}

\begin{IEEEkeywords}
energy modelling, humanoid robots, motor power, parameter identification,
constrained least squares, Unitree G1
\end{IEEEkeywords}

\section{Introduction}
\IEEEPARstart{H}{umanoid} robots are increasingly being deployed in
applications such as logistics, manufacturing, and agriculture,
where operation is constrained by finite onboard battery capacity.
Accurate prediction of electrical power consumption is therefore
important for energy-aware motion planning, mission-duration
estimation, battery management, and thermal protection of the
actuators. Among the major energy consumers of a humanoid platform,
the upper limbs can account for a significant fraction of the total
power budget during manipulation-intensive tasks.

Existing approaches to robot energy modelling range from purely
data-driven methods to detailed electro-mechanical models
\cite{lin2024bnlstm,chang2025hybrid,clochiatti2024ur5e}.
Data-driven methods can achieve high predictive accuracy but often
require substantial training data and may exhibit limited
extrapolation capability outside the operating conditions represented
in the training dataset \cite{lin2024bnlstm,chang2024unknownload}.
Conversely, first-principles models typically rely on detailed
knowledge of actuator, gearbox, and motor parameters that are often
unavailable for commercial robotic platforms
\cite{clochiatti2024ur5e,chang2025hybrid}.
Physics-based models that are linear in parameters provide an
attractive compromise, combining physical interpretability with
computationally efficient parameter identification through linear or
constrained least-squares estimation \cite{khalil2004,atkeson1986}.

Most existing studies focus on fixed-base industrial manipulators.
To the best of our knowledge, no physics-based electrical power
consumption model has been identified for a modern humanoid robot arm
using onboard power measurements as the regression target.
Furthermore, existing formulations rarely account for changes in
gravity-compensation power associated with different arm postures or
for coupling effects arising during simultaneous multi-joint motion.

This letter makes the following contributions:
\begin{itemize}
\item A physics-based linear-in-parameters electrical power model for
the seven joints of the Unitree G1 humanoid arm.

\item A baseline-torque correction that accurately captures changes
in copper losses relative to the static gravity-compensation load,
allowing the model to predict negative net power trajectories.

\item Pairwise joint-speed interaction terms that capture power
coupling during coordinated multi-joint motion.

\item A parameter-identification methodology based on trajectory-
averaged measurements that enables robust estimation from low-rate
onboard power sensors.

\item Experimental validation on 943 trajectories collected on a
physical Unitree G1 robot, achieving $R^2 = 0.933$ on the
identification dataset and $R^2 = 0.965$ on unseen validation
trajectories.
\end{itemize}

\section{Related Work}

Accurate modelling of robotic energy consumption has received increasing attention due to its importance for energy-aware motion planning, battery management, and sustainable robotic operation. Existing approaches can be broadly classified into physics-based models derived from robot dynamics and actuator characteristics, and data-driven methods that learn power-consumption patterns directly from experimental data.

Physics-based approaches seek to establish explicit relationships between robot motion, actuator torques, and electrical power consumption. Clochiatti \emph{et al.}~\cite{clochiatti2024ur5e} presented one of the most comprehensive electro-mechanical models for a collaborative manipulator, identifying inertial, friction, gearbox, and electrical parameters of the UR5e robot and estimating energy consumption directly from the robot dynamics. Similar efforts have employed analytical formulations combining mechanical power, motor efficiency, and actuator losses to estimate the energy requirements of industrial robots during task execution~\cite{hosseini2024trajectory,fabris2024parallel}. These approaches provide physically interpretable models and can support optimisation-based trajectory planning, but typically require access to detailed actuator specifications or manufacturer-provided parameters that are often unavailable.

To overcome this limitation, several recent studies have adopted data-driven or hybrid modelling techniques. Chang \emph{et al.}~\cite{chang2025hybrid} proposed a mechanism--data hybrid framework capable of predicting energy consumption without requiring prior knowledge of dynamic or electrical parameters. Purely data-driven approaches based on neural networks and recurrent architectures have also been reported for industrial manipulators~\cite{lin2024bnlstm,chang2024unknownload,vodovozov2024delta}. While these methods often achieve high predictive accuracy, their black-box nature limits physical interpretability and makes extrapolation outside the training distribution difficult. Recent benchmarking studies have highlighted the trade-off between predictive accuracy and model interpretability in robotic power estimation~\cite{lee2025framework}.

While recent advances in humanoid robotics have focused on actuator development~\cite{chignoli2021mit,sunbeam2025humanlevel,HUA2026103461,SUN2022104701,9895260}, whole-body control~\cite{9354897, 8294205}, and dynamic locomotion~\cite{Kong20183815, 7845678, 10.1007/978-981-16-0550-5_36, 8294205, 8260563, 11388961}, comparatively little attention has been devoted to modelling the electrical power consumption of humanoid manipulators. Existing energy-consumption studies remain largely centred on industrial robot arms, leaving open
questions regarding the applicability of these models to battery-powered humanoid platforms operating under different actuation and power-system constraints.

Overall, existing research has demonstrated the effectiveness of both
physics-based and data-driven approaches for estimating robotic energy
consumption. However, the literature remains predominantly focused on
industrial manipulators, with comparatively limited attention given to
electrical power modelling for humanoid robot arms.

\section{System Description}
\label{sec:system}

\subsection{Unitree G1 Left Arm}

The Unitree~G1 left arm has seven degrees of freedom: shoulder
pitch/roll/yaw, elbow, and wrist roll/pitch/yaw (SDK motor indices 15--21).
Each joint is driven by a BLDC motor paired with a harmonic-drive reducer.
Joint states (position~$q_i$, velocity~$\dq_i$, estimated torque~$\hat{\tau}_i$) are published at 100 (Hz) and available through ROS2 topics.

\subsection{Power Measurement}

Two sensors measure electrical power consumption in the Unitree G1.

\noindent\textbf{Battery Management System (BMS)}: provides pack voltage
$V_{\text{bms}}$ and discharge current $I_{\text{bms}}$ at approximately
1 (Hz). It measures total robot power
(135 (W) at rest), integrating legs, computing boards, and
all arm joints.

\noindent\textbf{Main-board sensor (MBS)}: provides an independent voltage and
current measurement at 1 (Hz). Its
120 (W) baseline is 15 (W) lower than the BMS. It monitors a dedicated arm/upper-body power rail, excluding most leg and
computer loads.  This makes it a lower-noise target for arm-specific modelling
and is used throughout this work.

\section{Per-Joint Power Model}
\label{sec:model}

\subsection{BLDC Motor and Gear Train}

Consider joint $i$ with winding resistance $R_i$, motor torque constant
$K_{m,i}$, gear ratio $n_i$, and gear efficiency $\eta_{g,i}$.  Neglecting
electrical transients, the motor obeys 
\begin{equation}
  V_i = R_i I_i + K_{m,i}\,\omega_{m,i},
  \label{eq:motor}
\end{equation}
where $\omega_{m,i} = n_i\,\dq_i$ and $I_i = \tau_i / (n_i\eta_{g,i}K_{m,i})$ ~\cite{siciliano2016springer}.

The total electrical power consumed by the motor is
\begin{equation}
  P_{\text{el},i} = V_i I_i
  = \underbrace{R_i I_i^2}_{P_{\text{Cu},i}}
  + \underbrace{\tau_{m,i}\,\omega_{m,i}}_{P_{\text{mech},m,i}}.
\end{equation}
Expressing everything at the joint level yields
\begin{equation}
  P_{\text{el},i}
  = \underbrace{\frac{1}{\eta_{g,i}}}_{a_i}\,\tau_i\dq_i
  + \underbrace{\frac{R_i}{(n_i\eta_{g,i}K_{m,i})^2}}_{b_i}\,\tau_i^2
  + c_i\,\lvert\dq_i\rvert
  + d_i\,\dq_i^2,
  \label{eq:joint_model}
\end{equation}
where the last two terms model Coulomb and viscous joint friction~\cite{olsson1998}.

\subsection{Baseline Copper-Loss Correction}
\label{sec:correction}

The net power $P_{\text{net},i} = P_{\text{el},i} - P_{\text{base},i}$
subtracts the \emph{static} contribution present when the arm rests at the
home posture ($\vect{q}=\vect{0}$, $\dq_i=0$).  At rest, the motor carries
the gravity-compensation current producing $b_i\tau_{0,i}^2$, which is
already included in $P_{\text{base}}$.  Substituting
$\Delta\tau_i^2(t) \coloneqq \tau_i^2(t) - \bar{\tau}_{0,i}^2$ into
\eqref{eq:joint_model} gives the \emph{net} per-joint model:
\begin{equation}
    P_{\text{net},i}
    = a_i\,\tau_i\dq_i
    + b_i\,\Delta\tau_i^2
    + c_i\,\lvert\dq_i\rvert
    + d_i\,\dq_i^2,
  \label{eq:net_model}
\end{equation}
where $\bar{\tau}_{0,i}^2$ is the mean squared torque during the static
pre-idle window.  $\Delta\tau_i^2$ is negative when the arm moves to a
configuration with lower gravity load than the home posture, which correctly
predicts $P_{\text{net},i}<0$ in such cases.

\subsection{Multi-Joint Interaction Term}

When several joints move simultaneously, the measured electrical
power exceeds the sum of the individual-joint contributions.
Similar effects have been reported in multi-axis servo systems
sharing a common DC bus, where interactions between drive
electronics, power flow, and energy exchange among axes can
influence the aggregate power demand
\cite{kaviani2012regenerative,hansen2014motion,chang2025hybrid}.  We augment
\eqref{eq:net_model} with pairwise interaction terms:
\begin{equation}
  P_{\text{net}}(t)
  = \sum_{i=1}^{7} P_{\text{net},i}(t)
  + \sum_{i<j} e_{ij}\,\lvert\dq_i\rvert\,\lvert\dq_j\rvert,
  \label{eq:full_model}
\end{equation}
where $e_{ij}\geq 0$.  The full parameter vector is
\begin{equation}
  \vect{\theta} =
  \bigl[\vect{a},\,\vect{b},\,\vect{c},\,\vect{d},\,\{e_{ij}\}\bigr]^\top
  \in\R^{49},
  \label{eq:theta}
\end{equation}
comprising 28 per-joint parameters (4 per joint $\times$ 7 joints) and
$\binom{7}{2}=21$ pairwise interaction coefficients.

\section{Parameter Estimation}
\label{sec:estimation}

\subsection{Linear-in-Parameters Form}

Model~\eqref{eq:full_model} is linear in $\vect{\theta}$:
\begin{equation}
  P_{\text{net},k} = \vect{\varphi}_k^\top \vect{\theta} + \varepsilon_k,
  \label{eq:linear}
\end{equation}
where the regressor $\vect{\varphi}_k\in\R^{49}$ at time $t_k$ stacks
$[\tau_i\dq_i,\,\Delta\tau_i^2,\,|\dq_i|,\,\dq_i^2,\,|\dq_i||\dq_j|]$
across all joint pairs.

\subsection{Outlier Filtering}
\label{sec:filtering}

Raw collected trajectories may be contaminated by two distinct artefacts that
inflate the regression residuals.

\noindent\textbf{Balance-compensation contamination.}
During fast or multi-joint manoeuvres the robot's locomotion stabiliser
adjusts leg posture to compensate for the arm's dynamic reaction, increasing
leg power consumption.  Because the MBS rail includes the upper body, this
leg energy leaks into the measured $P_\text{net}$.  We detect these episodes
by computing the IMU angular-velocity standard deviation,
$\sigma_\omega = \text{std}(\|\vect{\omega}_\text{body}\|)$, and reject any
trajectory for which $\sigma_\omega > 2.5\,\text{median}(\sigma_\omega)$.
In our dataset 104 of 1\,017 trajectories are rejected by this criterion.

\noindent\textbf{Statistical outliers.}
Residual sensor noise and infrequent hardware events produce a small number
of trajectories whose residuals fall far from the Gaussian bulk.  We apply
two passes of $3\sigma$ residual rejection: fit the model, identify trajectories
with $|\hat{r}_j| > 3\hat{\sigma}$, remove them, and refit.  This discards
an additional 16 trajectories and improves residual kurtosis from $77$ to
$2.97$ (near Gaussian).
After filtering, $M = 897$ trajectories are retained for the final fit.

\subsection{Trajectory-Level Aggregation}
\label{sec:aggregation}

The MBS sensor updates at $\approx\SI{1}{\hertz}$ while kinematics are
recorded at \SI{100}{\hertz}: each power reading is replicated
$\sim$\!\!100 times in the per-sample dataset.  Fitting on per-sample rows
gives a condition number $\kappa(\mat{\Phi}^\top\mat{\Phi}) \approx
7.8\times10^{11}$ and an $R^2 < 0.04$.

We instead aggregate each of the $M = 897$ filtered trajectories to a single
row:
\begin{equation}
  \bar{\vect{\varphi}}_j = \tfrac{1}{N_j}\sum_{k\in\mathcal{T}_j}\vect{\varphi}_k,
  \qquad
  \bar{P}_{\text{net},j} = \tfrac{1}{N_j}\sum_{k\in\mathcal{T}_j} P_{\text{net},k},
\end{equation}
where $\mathcal{T}_j$ is the index set for trajectory $j$ (median $N_j\approx 1\,016$
samples).  This averaging reduces noise by $\sqrt{N_j}\approx 32\times$ and
restores independence across observations.

\subsection{Constrained Optimisation}

The parameter vector is estimated by solving
\begin{equation}
  \vect{\theta}^* = \argmin_{\vect{\theta}}\;
  \bigl\|\bar{\mat{\Phi}}\vect{\theta} - \bar{\vect{y}}\bigr\|_2^2
  \label{eq:opt}
\end{equation}
subject to $a_i\geq 1$, $b_i\geq 0$, $c_i\geq 0$, $d_i\geq 0$,
$e_{ij}\geq 0$, where $\bar{\mat{\Phi}}\in\R^{897\times49}$ collects
the trajectory-averaged regressors and $\bar{\vect{y}}\in\R^{897}$ collects
the averaged net powers.  Problem~\eqref{eq:opt} is a strictly convex
Quadratic Programme (QP); it is solved using the interior-point solver
IPOPT~\cite{wachter2006} via the CasADi symbolic framework~\cite{andersson2019},
which enforces all inequality constraints exactly.

\section{Experimental Setup}
\label{sec:experiment}

\subsection{Recording Protocol}
All experiments are conducted with the robot standing, controlled by the high-level locomotion client, and using a safety harness and crane.; the right arm hangs at rest. The Dex3-1 hands are mounted on the left arm. Each trial follows four phases:

\begin{enumerate}
  \item \textbf{Pre-idle} (\SI{2}{\second}): arm held at
        $\vect{q}=\vect{0}$; used to estimate $P_{\text{base}}$ and
        $\bar{\tau}_{0,i}^2$.
  \item \textbf{Trajectory}: joint references sent at \SI{100}{\hertz}
        using cubic time-scaling~\eqref{eq:smoothstep}.
  \item \textbf{Post-idle} (\SI{2}{\second}): arm held at final posture.
  \item \textbf{Return}: smooth move back to $\vect{q}=\vect{0}$.
\end{enumerate}

Position and velocity gains are $k_p = \SI{400}{\newton\metre\per\radian}$
and $k_d = \SI{12}{\newton\metre\second\per\radian}$.

\subsection{Trajectory Set}

The dataset covers $N_\text{traj} = 1\,017$ distinct trajectories collected
in a single recording session ($M = 897$ retained after filtering).
Trajectories include:
(i)~single-joint sweeps for all 7 arm joints (shoulder pitch/roll/yaw,
elbow, wrist roll/pitch/yaw); (ii)~all two-joint combinations including
wrist pairs; (iii)~three-, four-, and higher-joint combinations; at five
speed levels
$\dq_\text{max} \in \{0.5, 1.0, 1.5, 2.0, 2.5\}$~\si{\radian\per\second}.
Including the wrist joints in the dataset is critical: without wrist-specific
trajectories the parameters $b_i$, $c_i$, $d_i$ for wrist joints are
unidentifiable and IPOPT assigns arbitrary values.

Joint positions follow cubic smooth-step scaling:
\begin{equation}
  s(\tau) = 3\tau^2 - 2\tau^3, \quad \tau\in[0,1],
  \label{eq:smoothstep}
\end{equation}
which ensures zero velocity at trajectory endpoints.

\subsection{Hold-Out Validation Trajectory Set}
\label{sec:val_set}

To assess generalisation beyond the training distribution, a separate set
of 46~trajectories is executed on the physical robot at four intermediate
speeds $\dq_\text{max} \in \{0.75, 1.25, 1.75, 2.25\}$~\si{\radian\per\second}
that are not present in the identification dataset.  Trajectories are rebuilt
from the \emph{exact same route waypoints} used during data collection---single-joint
sweeps for shoulder pitch/roll/yaw, elbow, and wrist pitch, plus wrist-only
two- and three-joint combinations---ensuring the same collision-free physical
paths are followed.  No additional filtering is applied to this set.

\section{Results}
\label{sec:results}

\subsection{Identified Parameters}

\begin{figure}[t]
  \centering
  \includegraphics[width=\columnwidth]{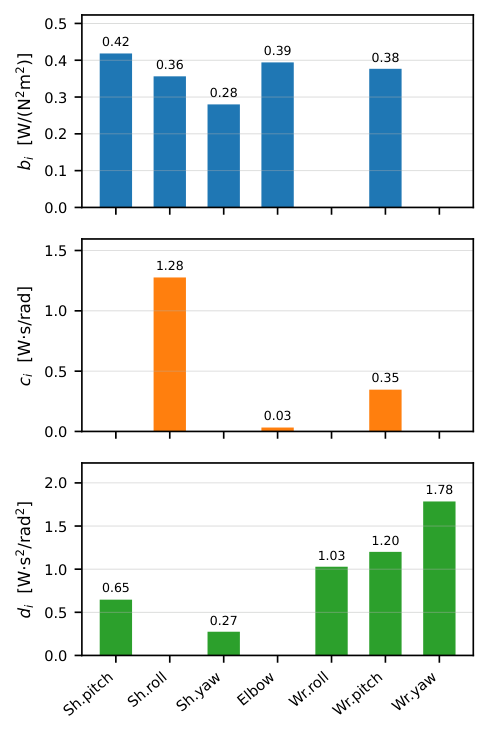}
  \caption{Identified power model coefficients for all 7 arm joints.
           Top: copper-loss coefficient $b_i$.
           Middle: Coulomb-friction coefficient $c_i$.
           Bottom: viscous-friction coefficient $d_i$.}
  \label{fig:params}
\end{figure}

Fig.~\ref{fig:params} shows the estimated coefficients $b_i$, $c_i$, $d_i$ for
all seven joints; Table~\ref{tab:params} lists the numerical values together
with the dominant loss mechanism for each joint, defined as
$\arg\max(b_i,c_i,d_i)$.  Viscous friction $d_i$ is the dominant term for four
of the seven joints---shoulder pitch and all three wrist joints---with wrist
yaw exhibiting the largest viscous coefficient overall
($d_6 = \SI{1.7848}{\watt\second\squared\per\radian\squared}$).  Copper losses
$b_i$ dominate shoulder yaw and the elbow ($b_2=0.2799$, $b_3=0.3942$).
Shoulder roll is the only joint for which Coulomb friction is the dominant
term, with $c_1=1.2767$ exceeding every $b_i$ and $d_i$ value in
Table~\ref{tab:params}; wrist pitch is the only other joint with a substantial
Coulomb contribution ($c_5=0.3464$), in addition to a non-negligible copper
term ($b_5=0.3768$), making it the joint with the most balanced split across
all three loss mechanisms.
The efficiency parameter $a_i=1$ for all joints (Table~\ref{tab:param_desc});
gear-train losses are absorbed into $b_i$, $c_i$, and $d_i$, a common outcome
when the harmonic drive operates in its high-efficiency regime.

Table~\ref{tab:eij} lists the full set of 21 pairwise interaction coefficients
$e_{ij}$.  The largest values occur between joint pairs that share a
kinematic-chain segment and are frequently actuated together: shoulder-roll /
elbow ($e_{1,3}=1.1053$), shoulder-pitch / elbow ($e_{0,3}=0.9090$),
shoulder-yaw / wrist-yaw ($e_{2,6}=0.8613$), shoulder-roll / wrist-yaw
($e_{1,6}=0.5111$), shoulder-pitch / wrist-yaw ($e_{0,6}=0.4365$), shoulder-yaw
/ elbow ($e_{2,3}=0.4054$), and wrist-roll / wrist-pitch ($e_{4,5}=0.3417$).

\begin{table}[t]
  \centering
  \caption{Identified Power Model Parameters (All 7 Joints)}
  \label{tab:params}
  \setlength{\tabcolsep}{4pt}
  \begin{tabular}{lccccc}
    \toprule
    Joint & $i$ & $b_i$ & $c_i$ & $d_i$ & Dominant term \\
    \midrule
    Sh.\ pitch & 0 & 0.4184 & 0.0000 & 0.6467 & viscous \\
    Sh.\ roll  & 1 & 0.3563 & 1.2767 & 0.0000 & Coulomb \\
    Sh.\ yaw   & 2 & 0.2799 & 0.0000 & 0.2749 & copper \\
    Elbow      & 3 & 0.3942 & 0.0324 & 0.0000 & copper \\
    Wr.\ roll  & 4 & 0.0000 & 0.0000 & 1.0282 & viscous \\
    Wr.\ pitch & 5 & 0.3768 & 0.3464 & 1.2004 & viscous \\
    Wr.\ yaw   & 6 & 0.0000 & 0.0000 & 1.7848 & viscous \\
    \bottomrule
  \end{tabular}

  \vspace{2pt}
  \footnotesize{$a_i=1$ for all joints. Units: $b_i$~[\si{\watt\per\newton\squared\metre\squared}],
  $c_i$~[\si{\watt\second\per\radian}], $d_i$~[\si{\watt\second\squared\per\radian\squared}].
  Dominant term $=\arg\max(b_i,c_i,d_i)$.  The full set of pairwise
  interaction coefficients $e_{ij}$ is given in Table~\ref{tab:eij}.}
\end{table}

\begin{table}[t]
  \centering
  \caption{Identified Pairwise Interaction Coefficients $e_{ij}$}
  \label{tab:eij}
  \setlength{\tabcolsep}{4pt}
  \begin{tabular}{cc|cc|cc}
    \toprule
    Pair & $e_{ij}$ & Pair & $e_{ij}$ & Pair & $e_{ij}$ \\
    \midrule
    (0,1) & 0.0000 & (1,3) & \textbf{1.1053} & (2,6) & \textbf{0.8613} \\
    (0,2) & 0.0000 & (1,4) & \textbf{0.1039} & (3,4) & 0.0000 \\
    (0,3) & \textbf{0.9090} & (1,5) & \textbf{0.1034} & (3,5) & 0.0000 \\
    (0,4) & 0.0000 & (1,6) & \textbf{0.5111} & (3,6) & 0.0591 \\
    (0,5) & 0.0000 & (2,3) & \textbf{0.4054} & (4,5) & \textbf{0.3417} \\
    (0,6) & \textbf{0.4365} & (2,4) & 0.0554 & (4,6) & 0.0000 \\
    (1,2) & 0.0000 & (2,5) & 0.0050 & (5,6) & 0.2936 \\
    \bottomrule
  \end{tabular}

  \vspace{2pt}
  \footnotesize{Joint indices: 0~shoulder pitch, 1~shoulder roll, 2~shoulder
  yaw, 3~elbow, 4~wrist roll, 5~wrist pitch, 6~wrist yaw.  Units:
  $e_{ij}$~[\si{\watt\second\squared\per\radian\squared}].  Bold:
  $e_{ij}>0.1$.}
\end{table}

\begin{table}[t]
  \centering
  \caption{Power Model Parameter Definitions and Final Identified Values}
  \label{tab:param_desc}
  \begin{tabular}{@{}p{0.09\columnwidth}p{0.63\columnwidth}p{0.20\columnwidth}@{}}
    \toprule
    Symbol & Description & Final value(s) \\
    \midrule
    $a_i$ & Inverse gear efficiency $1/\eta_{g,i}$ (Eq.~\eqref{eq:joint_model});
            scales mechanical power $\tau_i\dq_i$ &
            $1.0000$, $\forall i$ \\
    $b_i$ & Copper-loss (winding-resistance) coefficient; scales the
            baseline-corrected squared torque $\Delta\tau_i^2$ &
            Table~\ref{tab:params} \\
    $c_i$ & Coulomb-friction coefficient; scales $|\dq_i|$ &
            Table~\ref{tab:params} \\
    $d_i$ & Viscous-friction coefficient; scales $\dq_i^2$ &
            Table~\ref{tab:params} \\
    $e_{ij}$ & Pairwise joint-speed interaction coefficient; scales
            $|\dq_i||\dq_j|$ for $i<j$ &
            Table~\ref{tab:eij} \\
    \bottomrule
  \end{tabular}

  \vspace{2pt}
  \footnotesize{All 49 parameters ($\vect{\theta}\in\R^{49}$,
  Eq.~\eqref{eq:theta}) are non-negative and were estimated jointly by solving
  the QP of Eq.~\eqref{eq:opt} on the $M=897$ trajectory-averaged
  measurements of \S\ref{sec:estimation}.  The values reported in
  Tables~\ref{tab:params} and~\ref{tab:eij} are the converged parameters used
  throughout Section~\ref{sec:results} and embedded in the energy model used
  for RL training (\S\ref{sec:conclusion}).}
\end{table}

\subsection{Fit Quality}

Table~\ref{tab:fit} summarises goodness-of-fit for the final 7-joint model.
As discussed in \S\ref{sec:aggregation}, per-sample fitting on the $M=897$
filtered trajectories is numerically degenerate ($R^2<0.04$,
$\kappa(\Phi^\top\Phi)\approx7.8\times10^{11}$); trajectory-level aggregation
is therefore essential, not merely beneficial.  After aggregation, the full
model (all terms, outlier-filtered) achieves $R^2 = 0.933$ with
$\text{RMSE} = \SI{1.07}{\watt}$ and $\text{MAE} = \SI{0.86}{\watt}$
(Table~\ref{tab:fit}, top row); residual skewness is 0.15 and kurtosis is
2.97, close to a Gaussian distribution, indicating the linear model captures
the dominant power mechanisms.  Hold-out validation on 46 trajectories at
four unseen intermediate speeds (Table~\ref{tab:fit}, bottom row;
\S\ref{sec:validation}) yields $R^2 = 0.965$, confirming that the model
generalises beyond the training distribution rather than overfitting to the
five training speed levels.

\begin{table}[t]
  \centering
  \caption{Goodness-of-Fit: Final 7-Joint Power Model}
  \label{tab:fit}
  \begin{tabular}{lccc}
    \toprule
    Configuration & $R^2$ & RMSE [W] & MAE [W] \\
    \midrule
    \multicolumn{4}{l}{\textit{7-joint full model (897 filtered traj.)}} \\
    \textbf{MBS, traj.\ avg.\ + filter + all terms}
                                       & \textbf{0.933} & \textbf{1.07} & \textbf{0.86} \\
    \midrule
    \multicolumn{4}{l}{\textit{Hold-out validation (46 traj., unseen speeds)}} \\
    MBS, identified model              & 0.965 & 3.58 & 2.33 \\
    \bottomrule
  \end{tabular}
\end{table}

\begin{figure}[t]
  \centering
  \includegraphics[width=\columnwidth]{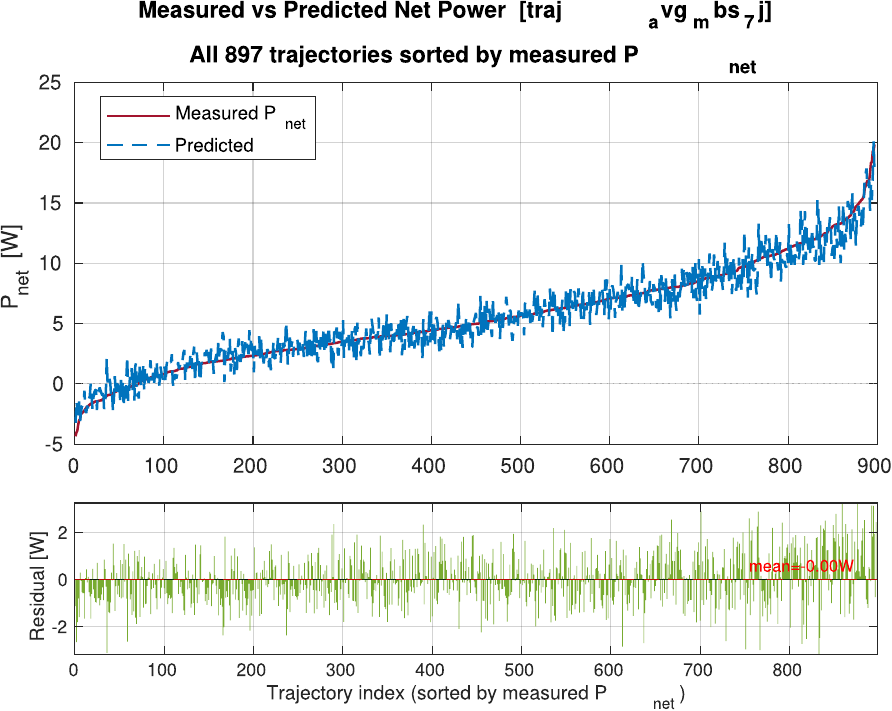}
  \caption{Measured (circles) and predicted (line) net power $P_\text{net}$
           for 897 trajectories sorted by measured value.
           Inset: residual distribution with fitted normal.
           $R^2=0.933$, RMSE$=\SI{1.07}{\watt}$.}
  \label{fig:fit}
\end{figure}

Fig.~\ref{fig:fit} shows predicted vs.\ measured $P_\text{net}$ for all 897
trajectories sorted by measured power.  The model tracks the full range
(\SI{-5}{\watt} to \SI{+27}{\watt}) with near-symmetric residuals; the
largest errors (up to \SI{3.3}{\watt}) occur in very-fast 4--5 joint
combined motions where simultaneous high-current draw may produce nonlinear
bus-voltage drop not captured by the linear interaction term.

\subsection{Power Term Contributions}

\begin{figure}[t]
  \centering
  \includegraphics[width=0.95\columnwidth]{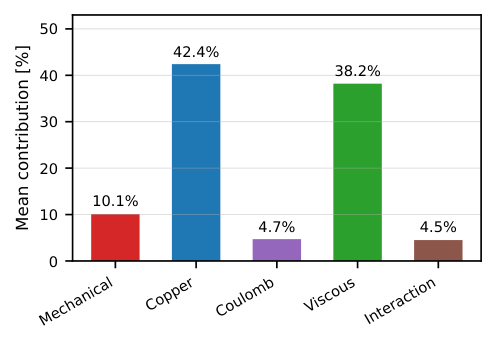}
  \caption{Mean power-term contributions (mechanical $a_i\tau_i\dq_i$,
           copper $b_i\Delta\tau_i^2$, Coulomb $c_i|\dq_i|$, viscous
           $d_i\dq_i^2$, and interaction $e_{ij}|\dq_i||\dq_j|$) evaluated on
           the 46-trajectory hold-out validation set using the identified
           parameters of Tables~\ref{tab:params} and~\ref{tab:eij}.  Copper
           losses and viscous friction dominate ($42.4\,\%$ and $38.2\,\%$,
           respectively).}
  \label{fig:breakdown}
\end{figure}

Fig.~\ref{fig:breakdown} shows the mean contribution of each power-model term,
evaluated on the 46-trajectory hold-out set (\S\ref{sec:validation}) using
the identified parameters of Tables~\ref{tab:params} and~\ref{tab:eij}.
Copper losses account for $42.4\,\%$ of total predicted power and viscous
friction for $38.2\,\%$---together explaining over $80\,\%$ of consumption,
consistent with the per-joint dominant terms identified in
Table~\ref{tab:params}.  Mechanical power contributes $10.1\,\%$, while
Coulomb friction ($4.7\,\%$) and the pairwise interaction terms ($4.5\,\%$)
are smaller but non-negligible, the latter justifying the inclusion of
$e_{ij}$ in the model.

\subsection{Hold-Out Validation}
\label{sec:validation}

Table~\ref{tab:fit} (bottom row) summarises the hold-out performance on the
46~trajectories described in \S\ref{sec:val_set}.  The identified model
achieves $R^2 = 0.965$, MAE~$= \SI{2.33}{\watt}$, and a bias of
\SI{-0.72}{\watt} (slight under-prediction).  The RMSE of \SI{3.58}{\watt}
is higher than the training RMSE of \SI{1.07}{\watt} because the validation
set contains trajectories with substantially higher power---the measured range
extends to \SI{121.7}{\watt} compared with the \SI{27}{\watt} ceiling of the
training set; on this wider range an RMSE of \SI{3.58}{\watt} corresponds to a
relative error of $2.9\,\%$ of the full scale.
The $R^2$ of $0.965$ exceeds the training-set value of $0.933$, confirming
that the model generalises to unseen intermediate speeds and does not overfit
to the five training speed levels.
The worst-case residual (\SI{17.2}{\watt}) occurs on a fast
elbow--wrist--pitch--yaw combination trajectory at \SI{1.0}{\radian\per\second};
this involves a simultaneous three-joint motion at a speed regime that was not
explicitly covered in the training set for that joint group.

\begin{figure}[t]
  \centering
  \includegraphics[width=\columnwidth]{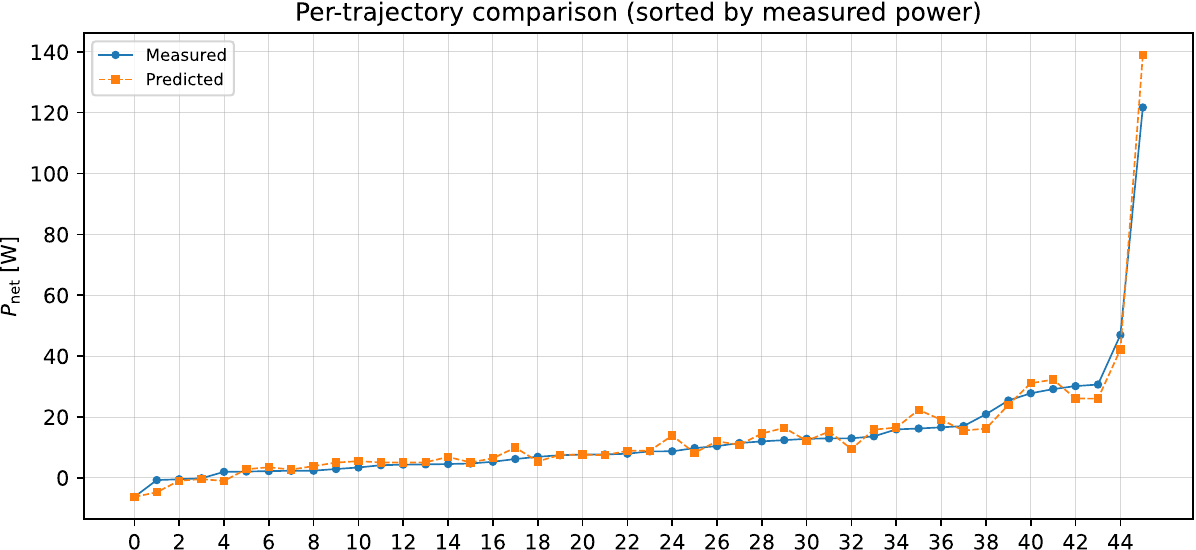}
  \caption{Predicted vs.\ measured net power $P_\text{net}$ for the
           46~hold-out trajectories (\S\ref{sec:val_set}), sorted by measured
           value.  The model tracks the measured power closely across the
           full \SI{121.7}{\watt} range; the single outlier (rightmost point)
           corresponds to the \SI{17.2}{\watt} worst-case residual on the
           fast elbow--wrist--pitch--yaw combination trajectory.}
  \label{fig:valpertraj}
\end{figure}

Fig.~\ref{fig:valpertraj} shows predicted vs.\ measured net power for all 46
hold-out trajectories, sorted by measured value.  Predictions track the
measured power closely across the full range, including the
near-zero/slightly-negative net power of the slowest trajectories.  The
single visible outlier is the highest-power trajectory (rightmost point),
where the model over-predicts by the \SI{17.2}{\watt} worst-case margin
discussed above; every other trajectory is reproduced to within a few watts.

\section{Discussion}
\label{sec:discussion}

\subsection{Effect of the Dex3-1 Hands}

All experiments were conducted with the Dex3-1 hands mounted, adding
rotational inertia at the wrist.  The large viscous-friction coefficients
identified for the wrist joints---wrist yaw ($d_6 = 1.7848$, the largest of
any joint) and wrist pitch ($d_5 = 1.2004$)---may partially reflect inertial
drag from the hand payload rather than pure gear friction.  A second
experimental round with the hands removed would isolate this contribution and
quantify how much of $d_5$ and $d_6$ is attributable to the Dex3-1 inertia
versus the bare wrist gear train; we leave this controlled comparison to
future work.

\subsection{Model Limitations}

\textbf{Temperature dependence.}  Motor winding resistance $R_i$ increases with
temperature as $R_i(T) = R_{0,i}[1 + \alpha_{\rm Cu}(T-T_0)]$.  Over long
experiments the effective $b_i$ drifts; a run-specific correction using the
measured motor temperature is left as future
work.

\textbf{Braking asymmetry.}  Harmonic drives are not back-drivable; during
deceleration the gear efficiency is $\eta_{g,i}$ (not $1/\eta_{g,i}$), making
the model asymmetric.  The current model uses a single $a_i$ for both modes.

\textbf{Collinearity.}  Trajectories where two joints follow identical velocity
profiles make $|\dq_i||\dq_j| \equiv \dq_i^2$, creating perfect collinearity
between viscous and interaction features.  Individual $d_i$ and $e_{ij}$ are
therefore not separately identifiable; only their combined effect is reliable.

\section{Conclusion}
\label{sec:conclusion}

We presented a physics-based, linear-in-parameters model for the electrical
power consumption of all 7 joints of a humanoid robot arm, achieving
$R^2 = 0.933$, RMSE~$= \SI{1.07}{\watt}$, MAE~$= \SI{0.86}{\watt}$ on
trajectory-averaged measurements from the on-board main-board power sensor.
Hold-out validation at four intermediate speeds not seen during identification
yields $R^2 = 0.965$, confirming that the model generalises beyond the
training distribution.
Key contributions are: (i)~a baseline copper-loss correction using the
pre-idle torque that correctly models negative net power in low-load
configurations; (ii)~pairwise joint-speed interaction terms for multi-joint
power coupling; (iii)~trajectory-level aggregation to eliminate the
\SI{1}{\hertz} sampling artefact; and (iv)~a two-stage outlier filter---IMU
stability rejection removing 104 of 1{,}017 trajectories, followed by
iterative $3\sigma$ residual rejection removing a further 16---that improves
residual kurtosis from 77 to 2.97, contributing to the overall identification
fit of $R^2=0.933$ on the remaining $M=897$ trajectories.  The model reveals
distinct loss-mechanism groupings across the arm (Table~\ref{tab:params}):
viscous friction dominates four of the seven joints (shoulder pitch and all
three wrist joints), copper losses dominate shoulder yaw and the elbow, and
shoulder roll is uniquely Coulomb-friction dominated---a fine-grained,
joint-specific picture invisible to torque-norm energy proxies.  The
identified model is embedded as a reward term in a deep RL training framework
for energy-efficient arm reaching, confirming its utility for downstream
motion planning tasks.

\bibliographystyle{IEEEtran}
\bibliography{biblio}

\end{document}